# A Dual-Contrastive Framework for Low-Resource Cross-Lingual Named Entity Recognition

Yingwen Fu
School of Information Science and Technology
Guangdong University of Foreign Studies
Guangzhou, China
fyinh@foxmail.com

Nankai Lin
School of Information Science and Technology
Guangdong University of Foreign Studies
Guangzhou, China
neakail@outlook.com

Ziyu Yang
School of Information Science and Technology
Guangdong University of Foreign Studies
Guangzhou, China
809241889@qq.com

Shengyi Jiang✉
School of Information Science and Technology
Guangdong University of Foreign Studies
Guangzhou, China
jiangshengyi@163.com

## ABSTRACT

Cross-lingual Named Entity Recognition (NER) has recently become a research hotspot because it can alleviate the data-hungry problem for low-resource languages. However, few researches have focused on the scenario where the source-language labeled data is also limited in some specific domains. A common approach for this scenario is to generate more training data through translation or generation-based data augmentation method. Unfortunately, we find that simply combining source-language data and the corresponding translation cannot fully exploit the translated data and the improvements obtained are somewhat limited. In this paper, we describe our novel dual-contrastive framework ConCNER for cross-lingual NER under the scenario of limited source-language labeled data. Specifically, based on the source-language samples and their translations, we design two contrastive objectives for cross-language NER at different grammatical levels, namely **Translation Contrastive Learning (TCL)** to close sentence representations between translated sentence pairs and **Label Contrastive Learning (LCL)** to close token representations within the same labels. Furthermore, we utilize knowledge distillation method where the NER model trained above is used as the teacher to train a student model on unlabeled target-language data to better fit the target language. We conduct extensive experiments on a wide variety of target languages, and the results demonstrate that ConCNER tends to outperform multiple baseline methods. For reproducibility, our code for this paper is available at https://github.com/GKLMIP/ConCNER.

## 1 Introduction

Named Entity Recognition (NER) [1] identifies the boundaries of entities in a text and correctly classifies them into predefined categories (e.g., locations, persons, organizations). Here is an example "Jenny Green comes from New York.", NER recognizes that the first two tokens "Jenny Green" refer to a person, while the last two tokens "New York" refer to an organization. As a fundamental task in Natural Language Processing (NLP), NER has multiple applications in various industrial products. For example, in commercial web search engines, such as Google, NER is crucial for Query Understanding [2], User Interest Modeling [3], and Question Answering [4]. For voice assistants such as Siri and Alexa, NER is a core component for Spoken Language Understanding (SLU) [5] and Dialogue System [6] Recently, deep neural networks have been widely applied in NER and have achieved good performance [7][8][9][10][11][12]. Unfortunately, deep neural network models usually require a large amount of training data, and thus neural NER models are successful for languages with massive labeled data. However, most languages, especially low-resource languages, do not have enough labeled data to train a fully supervised model. As the NER task is a representative token classification task, annotating a large number of samples could be expensive, time-consuming, and prone to human errors. The data-hungry problem becomes the biggest bottleneck for the application of neural networks to low-resource languages.

To reduce the cost of human annotating training data, cross-lingual NER has recently become a research hotspot. This technique aims to transfer knowledge from a source language

---

(usually a high-resource language with rich labeled data) to a target language (usually a low-resource language with little or even no labeled data). This paper focuses on zero-resource cross-lingual transfer, where **no labeled data is available in the target language** [13][14]. The recent methods for cross-lingual NER mainly fall into three categories: 1) Data transfer-based methods, which aims to build a pseudo labeled dataset for the target language by projecting alignment information [15][16][17][18][19] from a source language, and the target NER model is then trained on this pseudo dataset; 2) Model transfer-based methods [13][20][21][22][23], which focuses on training a shared NER model on the source labeled data with language-independent features and then applying the model directly to the target language; and 3) knowledge distillation-based models [24][25][26][27] based on teacher-student training framework are recently proposed to leverage large unlabeled data on the target language to further boost the cross-lingual NER performance.

These recent methods have shown promising performance in cross-lingual NER. However, most methods assume that there is abundant training data in the source language. When the training data size is reduced, the performances of methods above may significantly decrease [31]. However, the size of labeled data in the source language is relatively small in many real-life specific domains. Therefore, scaling out NER to a large number of target languages within a small number of source-language labeled data remains a grand challenge to the industry.

**Table 1: The F1-score performance comparison between NER performance in 1000 English samples and combination of 1000 English samples on WikiAnn dataset [48] and its translation on the target language. Following [13], we fine-tune mBERT with the data and directly test on the target languages.**

| Languages | Bulgarian | German | Hungarian |
|---|---|---|---|
| **En** | 67.06 | 67.35 | 52.03 |
| **En + Trans** | 66.98 | 66.15 | 47.92 |

To alleviate this issue, a common strategy is to combine source and target-language data (usually generated by translation or generation-based data augmentation methods) to further improve cross-lingual NER performance in low source-language labeled data scenarios using multilingual pre-trained language models (PLMs) [12][28] as the bridge. Unfortunately, as shown in Table 1, we find that simply combining source-language data and the corresponding translation cannot fully exploit the translated data and the improvement obtained is somewhat limited or even no performance gain is made. To tackle this, in this paper, we describe a novel approach to cross-lingual NER under the setting of a small amount of source-language data. We name our model ConCNER, a Contrastive Framework for Cross-lingual NER. Specifically, inspired by contrastive learning [29], two contrastive objectives are proposed to improve the cross-lingual NER performance, namely label contrastive objective and a translation contrastive objective, which exploit the source-language data and its translation in a more advanced way than just combing them together. The insight of the usage of contrastive learning on cross-lingual NER is to enlarge the representation gap between different labels to encourage token classification more correctly and close the representation gap between different languages to make transfer learning more possible. On one hand, the label contrastive objective encourages token representations with the same label to be closer and those with different labels to be more distant. This objective serves two purposes: 1) mitigate the negative effects of noisy translations to some extent and 2) through the contrastive learning of combined source and target-language data, the vector space of the two languages can be implicitly brought closer and more language-independent knowledge can be learned. On the other hand, the translation contrastive objective encourages the translated sentence representations to be closer and the other sentences representations to be more distant. This objective further explicitly brings the vector space of the two languages closer, reducing the damage of language differences on cross-linguistic NER performance.

Later, to better fit the target language, we extend knowledge distillation [30] to further exploit the language-specific language from a large amount of unlabeled target-language data. Concretely, we treat the proposed model above as the teacher model for predicting the probability distributions on the unlabeled target data. Then, a student model is trained with such soft-labeled data. The distilled student model can thus utilize both the task-specific knowledge from the teacher model and the language-specific knowledge from the unlabeled target language data.

In summary, we make the following contributions in this paper:

(1) We propose a contrastive framework for cross-lingual NER with two different contrastive objectives. To the best of our knowledge, this is the first to apply contrastive learning to cross-language NER.

(2) Instead of simply adding training data, we combine model transfer and data transfer methods in a more advanced way. Through contrastive learning, the usage of translation target data can be fully utilized.

(3) We extend knowledge distillation to the proposed model and combine it with contrastive learning to further improve the performance of the proposed model.

(4) With only a small amount of source language data, our approach achieves significant improvement on the cross-lingual NER task. When further incorporating with knowledge distillation, our approach achieves new SOTA performance. We also show the intuitive analysis of the contrastive representations.

The rest of the paper is organized as follows. We review the related work in Section 2 and present the proposed method in Section 3. We report an empirical experiment and analysis in Section 4 and draw a conclusion in Section 5.

## 2 Related Work

### 2.1 Cross-lingual NER

A Dual-Contrastive Framework for Low-Resource Cross-Lingual
Named Entity Recognition

Cross-lingual NER methods can be roughly categorized into data transfer-based, model transfer-based, and knowledge distillation-based methods.

Data transfer-based approach [15][16][17][18][19][31] aims to build a pseudo labeled dataset for a target language by projecting alignment information such as parallel resources [15][16] and machine translation [17][18][19][31] from a source language. Then, the target NER model is trained on this pseudo labeled dataset. In alignment methods, Wang and Manning [15] proposed to use bilingual parallel corpora to project model expectations from a high-resource language to a low-resource language. Ni, Dinu, and Florian [16] designed two co-decoding strategies to combine two projection methods: annotation projection on comparable corpora and projection of word embeddings between source and target languages. As for machine translation-based approaches, Mayhew, Tsai, and Roth [17] translated supervised data in a high-resource language into a low-resource language with a bilingual lexicon at phrase level. Xie et al. [18] translated the source language labeled data at the word level to generate pairwise labeled data for the target language. Jain, Paranjape, and Lipton [19] and Liu et al. [31] leveraged machine translation in a more advanced way that maintained sentence-level label sequences for translated data in the target language. The translation method proposed by Liu et al. [31] have recently achieved state-of-the-art (SOTA) performance in data transfer-based method so **we utilize it to translate source-language data in this paper.**

Model transfer-based approach [20][21][22][23][32] focuses on training a shared NER model on the source labeled data with language-independent features. This model is then directly applied in the target language. Specifically, Zirikly and Hagiwara [32] used graph propagation to generate multilingual gazetteers and cross-lingual word representation mappings. Wang, Peng, and Duh [20] and Yang, Salakhutdinov, and Cohen [21] leveraged multi-task learning to conduct the model transfer. They respectively introduced different objectives to learn a shared vector space between language pairs as well as NER task in the source language and then directly applied the model to the target language. Keung, Lu, and Bhardwaj [22] and Bari, Joty, and Jwalapuram [23] extended adversarial learning to model transfer-based cross-lingual NER. Recently, multilingual PLMs are adopted as the base model and transfer the knowledge in the supervised data of the source language to target languages [13][33][34]. These methods, taking advantage of large-scale unsupervised pre-training in multiple languages, achieve effective results in cross-lingual NER. In this paper, we introduce two contrastive objectives between source-language and its corresponding translated data as auxiliary objectives to learn a shared vector space between the source and target languages based on multilingual PLMs.

Recently, some works [24][25][26][27] extend knowledge distillation in cross-lingual NER based on a teacher-student training framework. The notion is to utilize a pre-trained source NER model to generate pseudo labels for target language data without external resources. Wu et al. [24] originally extended KD to cross-lingual NER within the teacher-student training framework. Later in order to reduce the impairment of some negative samples due to teacher model prediction errors, a few studies proposed some strategies to improve the standard KD process such as the combination of data transfer [25], adversarial discriminator [26], and reinforcement learning (RL) [27]. In this paper, to further boost the cross-lingual NER performance, we extend knowledge distillation which treats the contrastive model as the teacher to train a target-language student model.

## 2.2 Contrastive Learning

Contrastive learning is a representation learning method, which has been widely used in multiple representations such as graph representations [35], visual representations [36][37], response representations [38][39], text representations [40]. The main idea is to learn a representation by contrasting positive and negative sample pairs. Specifically, it puts positive pairs together and pushes negative pairs away. It can be classified into two groups. Self-supervised contrastive learning is often used for model pre-training. These methods propose different strategies to construct positive instances for each training sample such as hidden dropout masks [41], sentence-level augmentation [42], back-translation [43], and so on. In this paper, machine translation is used to produce positive instances for self-supervised contrastive learning which can better build a shared vector space between language pairs.

Recently, supervised contrastive learning is proposed and become popular. It builds positive pairs between instances with the same class label and puts their representations together. It is now widely applied to multiple tasks such as Natural Language Understanding (NLU) [44], Aspect-based Sentiment Analysis (ABSA) [45], Intent classification (IC) and slot filling (SF) [46]. This paper extends supervised contrastive learning to cross-lingual NER that can mitigate the negative effects of noisy translations to some extent and implicitly put the vector space of the two languages together.

## 3 Methodology

### 3.1 Problem Definition and Preliminaries

Cross-lingual named entity recognition is modeled as a sequence labeling problem. Given a sentence $w = \{w_0, w_1, \ldots, w_L\}$ with $L$ tokens as input, a NER model outputs a sequence of labels $= \{y_0, y_1, \ldots, y_L\}$, where $y_i \in Y = \{B-PER, B-LOC, B-ORG, B-MISC, I-PER, I-LOC, I-ORG, I-MISC, O\}$ indicates the label of the entity (or not an entity) of the corresponding token $w_i$. $D_{SRC} = \{(w, y)\}$ denotes a small-size labeled dataset of the source language. $B, I$ respectively represent the beginning, non-beginning of an entity. As for the target language, there is no labeled data for training but a test set $D_{Test}$ with labels for evaluation. Besides, we translate $D_{SRC}$ to the target language and construct a translated dataset $D_{TGT} = \{(w, y)\}$. $D_U = \{w\}$ denotes a large unlabeled data of the target language, which is used for knowledge distillation. We aim to learn a cross-lingual NER model by leveraging $D_{SRC}$, $D_{TGT}$, and $D_U$ to improve NER performance.

As shown in Figure 1, in this paper, we propose to leverage contrastive learning to make good use of $D_{TGT}$ for the cross-lingual NER task. The model $\mathcal{M}_1$ has three main components:

- Encoder that maps the input tokens into contextualized representations.
- Token classifier that calculates the label probability distribution of the input tokens and trained in a supervised manner with cross-entropy (CE) loss.
- Contrastive learning module that consists of two contrastive objectives: LCL that maximizes the agreement between the token representations with the same label while keeping them distant from other token representations in the same mini-batch and TCL that maximizes the agreement between one representation and its corresponding version that is translated from the same sentence while keeping it distant from other sentence representations in the same mini-batch.

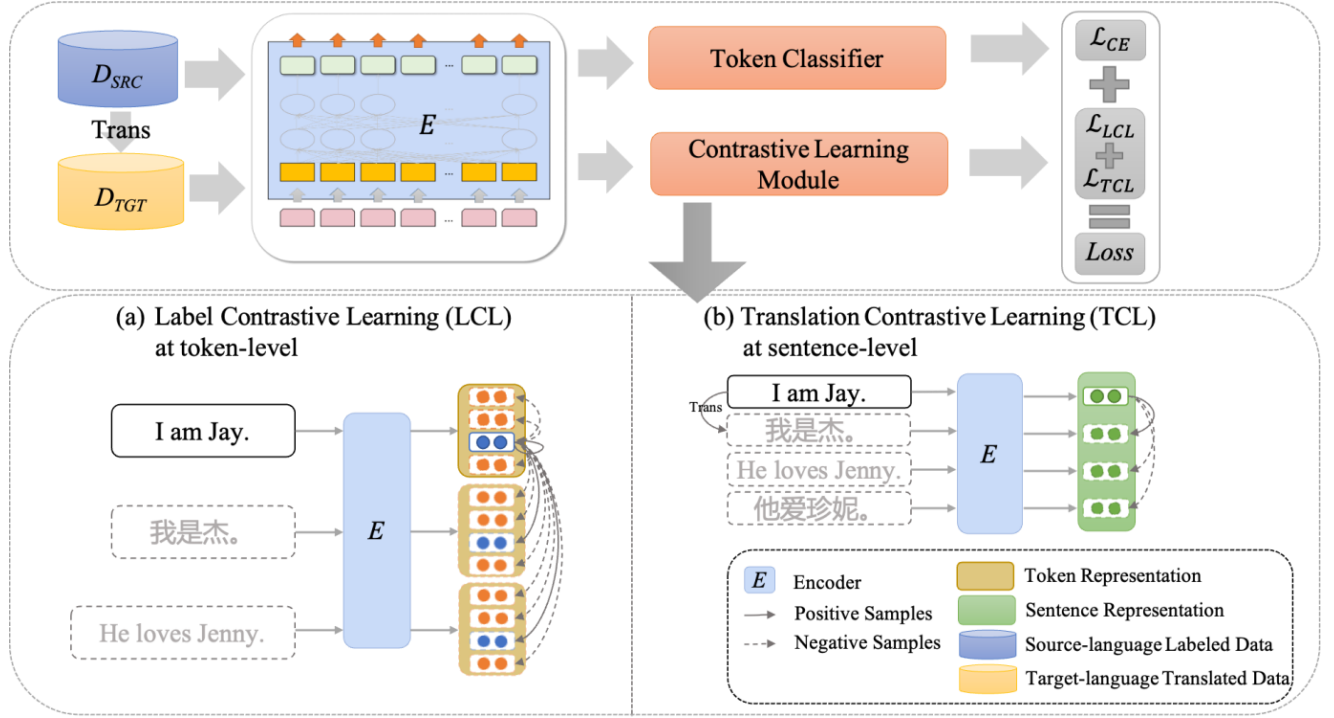

After that, we leverage knowledge distillation method to further improve the target-language NER performance. We follow the knowledge distillation process in [24] that train the student model to mimic the output probability distribution of entity labels by the teacher model, on the unlabeled data in the target language $D_{Tgt}$. Notably, the teacher model is $\mathcal{M}_1$ trained above.

## 3.2 Contrastive Model

*3.2.1 Encoder.* The adopted NER model consists of a transformer-based encoder layer. Specifically, given a sentence $\mathbf{w} = \{w_0, w_1, ..., w_L\}$ with $L$ tokens as input, the encoder layer $f_\theta$ maps it into a sequence of hidden vectors $\mathbf{h} = \{h_0, h_1, ..., h_L\}$:

**Figure 1:** The architecture of our contrastive framework for cross-lingual NER. The framework consists of an encoder, a token classifier, and a contrastive learning module. The contrastive learning module has two contrastive objectives: (a) LCL at the token level and (b) TCL at the sentence level.

$$\mathbf{h} = f_\theta(\mathbf{w}) \quad (1)$$

In general, any encoding model $f_\theta(\cdot)$ that produces a hidden state $h_i$ for the corresponding input token $w_i$ can be employed.

*3.2.2 Token Classifier.* Then, a multi-layer perceptron (MLP) classifier followed by the softmax function is leveraged to calculate the label probability distribution of $w_i$:

$$p(w_i, \Theta) = softmax(W \cdot h_i + b) \quad (2)$$

where $W$ and $b$ are the weight and bias term which are learnable parameters. Then the model is trained in a supervised manner by minimizing the CE loss of predicted probability $p(w_i, \Theta)$ and the ground truth label $y_i$. We train the model using the combination of the bilingual data, i.e., $D = D_{SRC} \cup D_{TGT}$.

$$\mathcal{L}_{CE} = \frac{1}{|D|} \sum_{(w,y) \in D} [-\frac{1}{L} \sum_{i=0}^{L} y_i \log p(w_i, \Theta)] \quad (3)$$

A Dual-Contrastive Framework for Low-Resource Cross-Lingual
Named Entity Recognition

*3.2.3 Contrastive Learning Module.* As stated above, our contrastive learning module is comprised of two objectives: LCL and TCL. During each training step, we randomly sample $N$ sentences from $D_{SRC}$ and their corresponding translations from $D_{TGT}$ to construct a mini-batch $\mathcal{B}$, resulting in $2N$ sentences. We assume that the mini-batch has $M$ tokens.

**LCL** is a token-level objective that aims to maximize the agreement between the token representations with the same label while keeping them distant from other token representations in the same mini-batch. Given a token $w_i$ where $0 < i < M$, its positive sample indexes are defined as $P = \{p: 0 < p < M \land p \neq i \land y_p = y_i\}$, and the label contrastive loss of $w_i$ is defined as:

$$\mathcal{L}_{LCL}^{w_i} = \frac{1}{|P|}\sum_{p \in P} -log \frac{exp\ (sim(h_i,h_p)/\tau_{LCL})}{\sum_{i=0}^{M} \mathbb{1}_{[k \neq i]} exp\ (sim(h_i,h_k)/\tau_{LCL})} \quad (4)$$

where $sim(\cdot)$ indicates the cosine similarity function, $\tau_{LCL}$ controls the temperature and $\mathbb{1}$ is the indicator. And the label contrastive loss of $\mathcal{B}$ is calculated by averaging $\mathcal{L}_{LCL}^{w_i}$ of all tokens:

$$\mathcal{L}_{LCL}^{\mathcal{B}} = \frac{1}{M}\sum_{i=0}^{M}\mathcal{L}_{LCL}^{w_i} \quad (5)$$

Finally, we average the $\mathcal{L}_{LCL}^{\mathcal{B}}$ of all mini-batches to obtain the final contrastive loss $\mathcal{L}_{LCL}$.

**TCL** is a sentence-level objective that aims to maximize the agreement between one representation and its corresponding version that is translated from the same sentence while keeping it distant from other sentence representations in the same mini-batch. We produce the sentence representations $S = \{r_i\}_{i=0}^{2N}$ through average pooling over the hidden states $\boldsymbol{h}$ from the encoder. Given a sentence presentation $r_i$, its positive sample is the sentence representation of its corresponding translation which is defined as $r_i^+$. The translation contrastive loss of $r_i$ is calculated as follows:

$$\mathcal{L}_{TCL}^{r_i} = -log \frac{exp\ (sim(r_i,r_i^+)/\tau_{TCL})}{\sum_{i=0}^{2N} \mathbb{1}_{[k \neq i]} exp\ (sim(r_i,r_k)/\tau_{TCL})} \quad (6)$$

And the translation contrastive loss of $\mathcal{B}$ is calculated by averaging $\mathcal{L}_{TCL}^{r_i}$ of all tokens:

$$\mathcal{L}_{TCL}^{\mathcal{B}} = \frac{1}{2N}\sum_{i=0}^{2N}\mathcal{L}_{TCL}^{r_i} \quad (7)$$

Finally, we average the $\mathcal{L}_{TCL}^{\mathcal{B}}$ of all mini-batches to obtain the final translation contrastive loss $\mathcal{L}_{TCL}$.

*3.2.4 Joint Training.* The three losses mentioned above are combined and jointly trained in this paper. For the overall pre-training loss $\mathcal{L}$, we sum up the CE loss and two contrastive losses, and $\alpha$ and $\beta$ are coefficients to balance the objectives:

$$\mathcal{L} = \alpha \mathcal{L}_{CE} + \beta(\mathcal{L}_{LCL} + \mathcal{L}_{TCL}) \quad (8)$$

After that, we can get a trained model $\mathcal{M}_1$ and use it as the teacher model for the following knowledge distillation process.

## 3.3 Distillation on Unlabeled Target Data

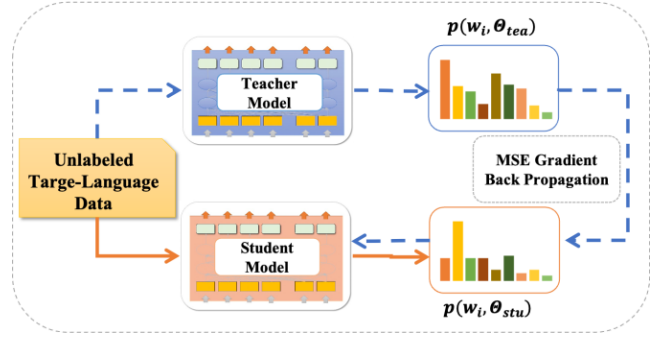

**Figure 2: The knowledge distillation process for cross-lingual NER.**

Knowledge distillation is considered to transfer knowledge from the source language to the target language as well as learning language-specific language from unsupervised target-language dataset $D_U$. As shown in Figure 2, after training the contrastive teacher model above, given a target-language sentence $\boldsymbol{w} = \{w_0, w_1, ..., w_L\}$ with $L$ tokens in $D_U$, we minimize the average mean square error (MSE) the predicted probability distributions of entity labels by the student model and those by the teacher model.

$$\mathcal{L}_{kd} = \frac{1}{|D_U|}\sum_{w \in D_U}[\frac{1}{L}\sum_{i=0}^{L} MSE(p(w_i, \Theta_{stu}), p(w_i, \Theta_{tea}))] \quad (9)$$

$\Theta_{stu}$ and $\Theta_{tea}$ are respectively the parameters of the student and teacher models and $\Theta_{tea}$ is fixed during the training; $p(\cdot, \Theta_{stu})$ and $p(\cdot, \Theta_{tea})$ are respectively the predicted distributions of the student and teacher models. The teacher and student models share the same architecture with different parameter weights. Notably, contrastive learning is not leveraged in the student model training.

## 3.4 Training Process and Inference

Algorithm 1 gives the overall training process of the proposed ConCNER framework in this paper.

| Algorithm 1: Training Process of ConCNER |
|---|
| 1: # Contrastive Teacher Model Training |
| 2: # $D_{SRC}$: Labeled NER data in the source language |
| 3: # $D_{TGT}$: Translated NER data in the target language |
| 4: initial the teacher model $\mathcal{M}_1$ with $mBERT$ |
| 5: **while** *not converge* **do** |
| 6:    **for** all $(\boldsymbol{w}, \boldsymbol{y})$ in $D_{src}$ and its translation $(\boldsymbol{w}^+, \boldsymbol{y}^+)$ in $D_{TGT}$ **do** |
| 7:       Update $\mathcal{M}_1$ with Equation (8); |
| 8:    **end for** |
| 9: **end while** |
| 10: |
| 11: |
| 12: # Knowledge Distillation Process |
| 13: # $D_U$: Unlabeled data in the target language |
| 14: initial the student model with $mBERT$ |
| 15: initial the teacher model with $\mathcal{M}_1$ |
| 16: **while** *not converge* **do** |
| 17:    **for** all $\boldsymbol{w}$ in $D_U$ **do** |
| 18:       Predict the soft labels for $\boldsymbol{w}$ with the teacher model; |
| 18:       Update the student model with Equations (9); |
| 19:    **end for** |
| 20: **end while** |

For inference on the target language, we only utilize the trained student model to predict the probability distribution of entity labels for each token $w_i$ in a test sentence $\boldsymbol{w}$. Then we take the entity label $c \in Y$ with the highest probability as the predicted label for $w_i$.

$$y_i = \arg\max_c p(w_i, \Theta_{stu})_c \quad (10)$$

## 4 Experiment

### 4.1 Datasets

**Table 2: Statistics of the datasets.**

**(a) CoNLL statistics.**

| Language | Type | Train | Dev | Test |
|---|---|---|---|---|
| **English-en** (CoNLL-2003) | Sentence | 14,987 | 3,466 | 3.684 |
| | Entity | 23,499 | 5,942 | 5,648 |
| **Spanish-es** (CoNLL-2002) | Sentence | 8,323 | 1,915 | 1,517 |
| | Entity | 18,798 | 4,351 | 3,558 |
| **Dutch-nl** (CoNLL-2002) | Sentence | 15,806 | 2,895 | 5,195 |
| | Entity | 13,344 | 2,616 | 3,941 |
| **German-de** (CoNLL-2003) | Sentence | 12,705 | 3,068 | 3,160 |
| | Entity | 11,851 | 4,833 | 3,673 |

**(b) WikiAnn statistics.**

| Language | Type | Train | Dev | Test |
|---|---|---|---|---|
| **English-en** | Sentence | 20,000 | 10,000 | 10,000 |
| | Entity | 27,931 | 14,146 | 13,958 |
| **Arabic-ar** | Sentence | 20,000 | 10,000 | 10,000 |
| | Entity | 22,500 | 11,266 | 11,259 |
| **Hindi-hi** | Sentence | 5,000 | 1,000 | 1,000 |
| | Entity | 6,124 | 1,226 | 1,228 |
| **Chinese-zh** | Sentence | 20,000 | 10,000 | 10,000 |
| | Entity | 25,031 | 12,493 | 12,532 |

The CoNLL2002 [1] and 2003 NER datasets [47] are used for evaluation which contain dataset in four different languages: English [en], German [de], Dutch [nl] and Spanish [es]. In addition, we select three non-western languages (Arabic [ar], Hindi [hi], and Chinese [zh]) from another multilingual NER dataset WikiAnn [48] to evaluate the generalization and scalability of the proposed framework. Each dataset is split into training, development, and test sets. We focus on 4 entity types (i.e., **PER, LOC, ORG, and MISC**) of all datasets. We convert the data to IOB2 format. The detailed statistics of these datasets are shown in Table 2. For both CoNLL and WikiAnn, we use English as the source language and the others as target languages. We follow the steps described in [31] to translate the English training data $D_{SRC}$ to the target languages based on Google translation system[1] to construct $D_{TGT}$. For target languages, we remove the entity labels in the corresponding training data and adopt them as the unlabeled target-language instances $D_U$ for the knowledge distillation process. To follow the zero-shot setting, we use the English development set to select the best checkpoints and evaluate them directly on the target-language test sets.

### 4.2 Implementation Details

We implement ConCNER based on HuggingFace's Transformers[2] with PyTorch[3] in one single NVIDIA Titan RTX GPU. The basic encoders for all variants are initialized with mBERT[4]. Each model has 12 Transformer layers, 12 self-attention heads and 768 hidden units. Epochs, batch size, maximum sequence length, learning rate, and optimizer are set as 30, 32, 128, 5e-5, and AdamW [49] respectively. Notably, the batch size indicates the source-language data size and represents $N$ in subsection 3.2. Each experiment is repeated 5 times and the average entity-level F1-score is reported. For the hyperparameter $\tau_{tcl}$ and $\tau_{tcl}$ respectively in Equation (4) and (6), we set them as 0.1 consistently. $\alpha$ and $\beta$ in Equation (8) are simply set as 0.5 and 0.25. The target models are evaluated on the source-language development set every 100 steps and the checkpoints are saved based on the evaluation results.

To simulate low-resource scenarios, we randomly sample 500, 1000, and 2000 sentences from the gold English training set and translate them into target languages. When evaluating CoNLL-2002/2003 datasets, we use CoNLL-2003 English dataset as the source-language data and when evaluating WikiAnn datasets, we use WikiAnn English dataset as the source-language data.

### 4.3 Baseline Methods

We compare our method with the following baselines. [13] is a representative model transfer-based method that is trained using monolingual source-language data and evaluated in the target language. [24] originally extends knowledge distillation to cross-lingual NER based on a teacher-student training framework. MulDA [31] proposes an advanced labeled sequence translation method to translate source-language training data to target languages and achieves SOTA performance in the previous data-based methods.

### 4.4 Main Results

Table 3 shows the results on the cross-lingual NER of our methods and the baseline methods under different sizes of source-language labeled data. In general, ConCNER outperforms the three baseline methods by a large margin and tends to establish

---
[1] https://cloud.google.com/translate
[2] https://github.com/huggingface/transformers
[3] https://pytorch.org/
[4] https://huggingface.co/bert-base-multilingual-cased

A Dual-Contrastive Framework for Low-Resource Cross-Lingual
Named Entity Recognition

the new SOTA performance under the scenario of limited source-language data. The results indicate the effectiveness of the proposed method in this paper.

It can be seen from the table that simply expanding the training set through translation from source to target language does not appear to achieve significant improvements. What is more, improvement gain decreases as the source-language data size increases. This suggests that simply adding translated data to the training set cannot fully exploit the translated data. Our method makes use of translated data in a more advanced way, with TCL that can minimize the distance between translated sentence pairs, allowing the model to learn the shared space between language allowing the model to learn the shared space between language pairs effectively. And applying LCL to bilingual data encourages the model to better distinguish the token representations of different labels. Furthermore, LCL can implicitly minimize the distance between the token representations of the same labels in different languages which may also help to learn the bilingual shared space. They can significantly boost the model performance.

Table 3: Main results of our method and the baseline models. Notes: All results are re-implemented. In MulDA baseline, for better comparison, we only report the performance of combining the source language data with its translation.

| En size | | es | nl | de | ar | hi | zh | Avg. |
|---|---|---|---|---|---|---|---|---|
| **500** | Wu and Dredze [13] | 66.66 | 70.14 | 59.33 | 37.02 | 57.06 | 43.63 | 55.64 |
| | Wu et al. [24] | 69.99 | 72.28 | 62.50 | 40.49 | 58.40 | 46.38 | 58.34 |
| | MulDA [31] | 67.36 | 72.05 | 62.55 | 37.55 | 58.29 | 44.84 | 57.11 |
| | ConCNER | **71.08** | **75.19** | **66.96** | **41.84** | **61.87** | **47.68** | **60.77** |
| **1000** | Wu and Dredze [13] | 69.34 | 70.84 | 63.15 | 41.83 | 57.78 | 44.42 | 57.89 |
| | Wu et al. [24] | 71.13 | 73.84 | 65.39 | 43.71 | 59.75 | 45.50 | 59.89 |
| | MulDA [31] | 69.93 | 71.61 | 65.29 | 38.70 | 60.16 | 45.64 | 58.56 |
| | ConCNER | **73.43** | **76.25** | **68.74** | **41.96** | **64.95** | **49.21** | **62.42** |
| **2000** | Wu and Dredze [13] | 72.87 | 75.89 | 68.36 | 44.37 | 59.91 | 45.56 | 61.16 |
| | Wu et al. [24] | 74.96 | 78.16 | 70.97 | 46.53 | 62.08 | 47.86 | 63.43 |
| | MulDA [31] | 72.49 | 75.24 | 68.85 | 42.45 | 61.58 | 47.03 | 61.27 |
| | ConCNER | **75.57** | **78.35** | **72.82** | **46.97** | **65.69** | **50.25** | **64.94** |

Besides, [24] achieves the SOTA performance in all baselines which contributes to the knowledge distilled from a teacher model to the target model on unlabeled target-language data. We find that the performance of the student model may largely depend on that of the teacher model. Our method can effectively improve the teacher performance and thus fits the target languages better.

By analyzing the results of different source-language data sizes, we can see that ConCNER can give better results than the baseline methods with fewer data (ConCNER at the data size of 500 achieves better results than the baseline methods at the data size of 1000, and this situation also occurs in the data size of 1000). This further verifies that contrastive learning is effective for the zero-shot cross-lingual NER task.

In summary, ConCNER is less data-consuming and more effective in scenarios where source-language training data is limited. This may make it convenient for industrial applications.

### 4.5 Further Analysis

*4.5.1 Ablation Study.* We conduct experiments on different variants of the proposed framework to investigate the contributions of different components. Table 4 presents the results of removing one component at a time. There are 4 different variants: 1) *w/o* knowledge distillation: we remove the knowledge distillation process and use the contrastive teacher model $\mathcal{M}_1$ as the final model; 2) *w/o* TCL: we remove the TCL objective, and the overall objective is calculated as $\mathcal{L} = \alpha\mathcal{L}_{CE} + \beta\mathcal{L}_{LCL}$ where $\alpha$ and $\beta$ are set as 0.5; 3) *w/o* LCL: we remove the TCL objective, and the overall objective is calculated as $\mathcal{L} = \alpha\mathcal{L}_{CE} + \beta\mathcal{L}_{TCL}$ where $\alpha$ and $\beta$ are set as 0.5; and 4) *w/o* TCL and LCL: we remove all contrastive objectives and train the teacher model on the combination of source-language data and its corresponding translation with the loss $\mathcal{L} = \mathcal{L}_{CE}$. All the variants are trained in the source-language data size of 1000.

Table 4 shows the results of the variants and ConCNER. We can see that generally all components contribute to the cross-lingual setting. The full model consistently achieves the best performance on all languages experimented. "*w/o* knowledge distillation" suffers from a decrease of 2.85% in F1-score on average. This demonstrates the effectiveness of knowledge distillation method because it can capture massive target-language knowledge from the unlabeled target-language data and encourage the model to better fit the target language. Both "*w/o* LCL" and "*w/o* TCL" settings outperform the "*w/o* TCL and LCL" setting, indicating that the proposed two contrastive objectives are efficient ways to improve the target-language NER performance.

The two contrastive objectives can fully exploit the correlation between the source-language data and the translated target-language data and enhance the consistency between the language pairs thus making more contributions that just simply combining two kinds of data. In addition, it seems that TCL contributes more that LCL with an average F1-score gain of 0.2%. This indicates that TCL is a more effective way to transfer knowledge across languages since it explicitly closes the representation gap between different languages to encourage transfer learning as much as possible. It can make the vector space more language independent.

*4.5.2 Effectiveness of contrastive learning.* To better demonstrate how contractive objectives affect the cross-lingual NER performance, we train the NER model on the following variants of three target languages (Dutch, Spanish, and German): 1) En: train the models on English data with the NER objective only; 2) Trans: train the models on the translated target data with the NER objective only; 3) En+LCL: train the models on English data with the NER and LCL objectives; 4) Trans+LCL: train the models on the translated target data with the NER and LCL objectives; 5) En+Trans: train the models on the combination of English and translated data with the NER objective; 6) En+Trans+LCL: train the models on the combination of English and translated data with the NER and LCL objectives; 7) En+Trans+LCL+TCL: train the models on the combination of English and translated data with the NER, LCL and TCL objectives. Notably, to explore the effect of comparative learning more intuitively, all the above variants do not incorporate the knowledge distillation process.

**Table 4: The results of ablation study. Notes: Due to the space limitation, all the variants are trained in the source-language data size of 1000.**

|  | es | nl | de | ar | hi | zh | Avg. |
| --- | --- | --- | --- | --- | --- | --- | --- |
| *w/o* **knowledge distillation** | 71.10 | 72.92 | 66.31 | 39.56 | 61.18 | 46.37 | 59.57 |
| *w/o* **TCL** | 72.60 | 75.29 | 67.89 | 41.18 | 64.02 | 48.37 | 61.56 |
| *w/o* **LCL** | 72.88 | 75.42 | 68.42 | 41.41 | 64.08 | 48.34 | 61.76 |
| *w/o* **TCL and LCL** | 72.19 | 74.27 | 66.93 | 40.87 | 63.72 | 48.03 | 61.00 |
| **ConCNER** | **73.43** | **76.25** | **68.74** | **41.96** | **64.95** | **49.21** | **62.42** |

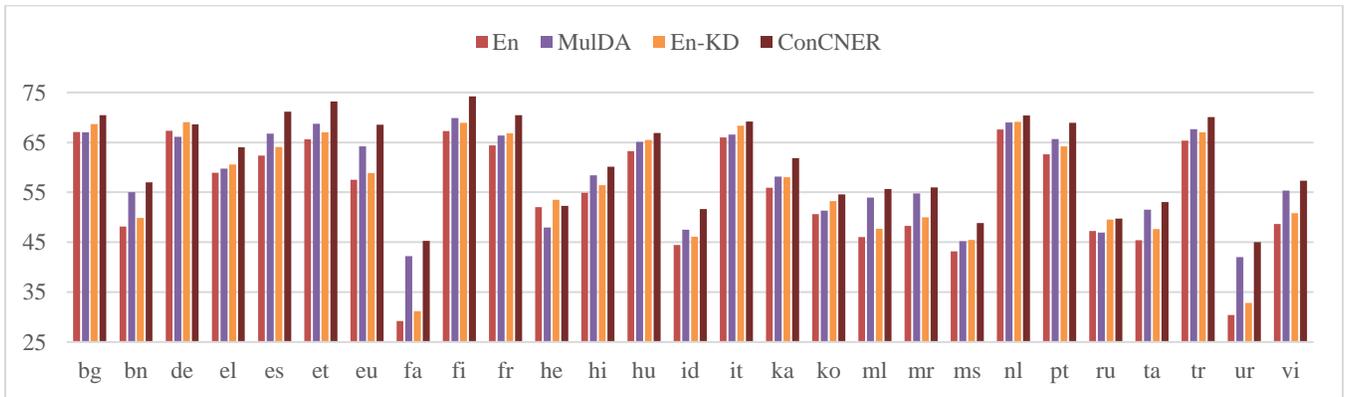

**Figure 3: The results of our method and the baseline models on more languages.**

The results are presented in Table 5. We can see that generally both LCL and TCL contribute to the cross-lingual setting. A special case is that applying LCL on English-only data (En+LCL) may negatively impair the model performance (Italicized data in Table 5). A possible reason is that through contrastive learning, some language-specific entity knowledge is extracted from the source-language data, which may not be applicable to the target language and therefore negatively optimized the target-language NER. In contrast, adopting LCL in bilingual data (En+Trans+LCL) can minimize the distance between token representations with the same label in different languages, thus implicitly pushing the representations of the language pair together. This facilitates the model to learn language-independent features and reduces the negative impact of source language-specific knowledge. The use of TCL (En+Trans+LCL+TCL) can explicitly bridge the gap between the language pair and create a bilingual shared vector space. It can further learn language-independent features and improve cross-lingual NER performance. En+Trans+LCL+TCL outperforms En by average F1-scores (3.05%, 2.33%, 1.35%) and En+Trans by average F1-scores (1.11%, 1.17%, 1.19%) in different data sizes of (500, 1000, 2000). These experimental

A Dual-Contrastive Framework for Low-Resource Cross-Lingual
Named Entity Recognition

results verify that the combination of LCL and TCL is capable of enhancing target-language NER by learning more language-independent entity-centric information.

*4.5.3 Evaluation on more languages.* We evaluate the proposed framework on a wider range of target languages in this section. We use more 27 languages in WikiAnn dataset as target languages and use English as the source language. 1000 English sentences are sampled from the gold data to simulate the low resource scenarios. Four models are evaluated: 1) En [13]; 2) MulDA [31]; 3) En-KD [24] and 4) ConCNER.

The results are summarized in Figure 3. It can be seen that ConCNER achieves the best results in the vast majority of target languages. Among them, ConNER can get a more significant boost in distant language pairs, such as bn, fa, id, ml, and so on, which should be credited to language-independent entity-centric knowledge obtained from labeled source-language data and its corresponding translation through two contrastive objectives and target-language knowledge obtained from unlabeled target-language data via knowledge distillation.

**Table 5: Effectiveness of contrastive learning.**

| En size | | es | nl | de | Avg. |
|---|---|---|---|---|---|
| 500 | En | 66.66 | 70.14 | 59.33 | 65.38 |
| | Trans | 65.38 | 71.53 | 62.39 | 66.43 |
| | En+Trans | 67.36 | 72.05 | 62.55 | 67.32 |
| | En+LCL | 68.10 | *69.79* | 59.59 | 65.83 |
| | Trans+LCL | 66.10 | 72.28 | 62.69 | 67.02 |
| | En+Trans+LCL | 67.84 | 72.40 | 63.38 | 67.87 |
| | En+Trans+LCL+TCL | **68.76** | **72.69** | **63.85** | **68.43** |
| 1000 | En | 69.34 | 70.84 | 63.15 | 67.78 |
| | Trans | 68.96 | 70.78 | 63.81 | 67.85 |
| | En+Trans | 69.93 | 71.61 | 65.29 | 68.94 |
| | En+LCL | 68.88 | *69.09* | *60.98* | 66.32 |
| | Trans+LCL | 68.34 | 72.09 | 64.60 | 68.34 |
| | En+Trans+LCL | 70.83 | 72.43 | 65.46 | 69.57 |
| | En+Trans+LCL + TCL | **71.10** | **72.92** | **66.31** | **70.11** |
| 2000 | En | 72.87 | 75.89 | 68.36 | 72.37 |
| | Trans | 71.53 | 75.02 | 67.63 | 71.39 |
| | En+Trans | 72.49 | 76.24 | 68.85 | 72.53 |
| | En+LCL | 73.82 | 76.14 | 67.94 | 72.63 |
| | Trans+LCL | 71.72 | 74.14 | 68.60 | 71.49 |
| | En+Trans+LCL | 72.97 | 76.46 | 68.85 | 72.76 |
| | En+Trans+LCL + TCL | **74.21** | **76.60** | **70.35** | **73.72** |

## 5 Conclusion

In this paper, we propose a dual-contrastive framework for cross-lingual named entity recognition (NER) with limited source-language labeled data. Based on the source-language samples and their translations, we design two contrastive objectives for cross-lingual NER at different grammatical levels, namely **T**ranslation **C**ontrastive **L**earning and **L**abel **C**ontrastive **L**earning (**LCL**). Furthermore, we leverage the available information from unlabeled target-language data via knowledge distillation to further improve the target-language NER performance. We report a series of experiments on several widely used datasets. The results indicate that the proposed framework tends to outperform the existing methods and establishes a competitive performance on the cross-lingual NER task with limited source-language labeled data. Besides, it also has great potential in multi-source settings and other cross-lingual tasks, which are left for future work.

One potential limitation of the proposed framework is that we treat each label as a separate unit in contrastive learning and ignore the relationships between different labels. For example, there may be a relationship between the "B-PER" and "I-PER" labels. This ignorance may be detrimental to the performance of contrastive learning, and we will explore the integration of label relationships in contrastive learning in future work.